\pdfoutput=1
\documentclass[11pt,letterpaper]{article}
\usepackage{naaclhlt2010}
\makeatletter
\newcommand{\@BIBLABEL}{\@emptybiblabel}
\newcommand{\@emptybiblabel}[1]{}
\makeatother
\usepackage{times}
\usepackage{latexsym}
\usepackage{fancyhdr} 
\usepackage{hyperref}

\makeatletter
\gdef \@makecol {%
   \ifvoid\footins
     \setbox\@outputbox \box\@cclv
   \else
     \setbox\@outputbox \vbox {%
       \boxmaxdepth \@maxdepth  
       \unvbox \@cclv
       \vskip \skip\footins
       \color@begingroup
         \normalcolor   
         \footnoterule  
         \moveright\paperwidth\rlap{%
           \kern-\paperwidth\box \footins
         }%
       \color@endgroup
       }%
   \fi   
   \let\@elt\relax
   \xdef\@freelist{\@freelist\@midlist}%
   \global \let \@midlist \@empty
   \@combinefloats
   \ifvbox\@kludgeins   
     \@makespecialcolbox
   \else
     \setbox\@outputbox \vbox to\@colht {%
       \@texttop
       \dimen@ \dp\@outputbox
       \unvbox \@outputbox   
       \vskip -\dimen@
       \@textbottom   
       }%
   \fi   
   \global \maxdepth \@maxdepth
}
\makeatother

\title{Using Mechanical Turk to Build Machine Translation Evaluation Sets}

\author{Michael Bloodgood\\
  Human Language Technology\\
  Center of Excellence\\
  Johns Hopkins University\\
  {\tt bloodgood@jhu.edu}  \And
  Chris Callison-Burch\\
  Center for Language and\\
  Speech Processing\\
  Johns Hopkins University\\
  {\tt  ccb@cs.jhu.edu}}

\date{}

\begin{document}

\thispagestyle{fancy}

\maketitle
\begin{abstract}
Building machine translation (MT) test sets is a relatively expensive task. 
As MT becomes increasingly desired for more and more language pairs and more and more domains, it becomes necessary to build test sets for each case. 
In this paper, we investigate using Amazon's Mechanical Turk (MTurk) to make MT test sets cheaply. 
We find that MTurk can be used to make test sets much cheaper than professionally-produced test sets.
More importantly, in experiments with multiple MT systems, we find that the MTurk-produced test sets yield essentially the same conclusions regarding system performance as 
the professionally-produced test sets yield. 
\end{abstract}

\section{Introduction} \label{introduction}

Machine translation (MT) research is empirically evaluated by comparing system output against reference human translations, typically using automatic evaluation metrics. 
One method for establishing a translation test set is to hold out part of the training set to be used for testing. 
However, this practice typically overestimates system quality when compared to evaluating on a test set drawn from a different domain. 
Therefore, it's necessary to make new test sets not only for new language pairs but also for new domains.

Creating reasonable sized test sets for new domains can be expensive. 
For example, the Workshop on Statistical Machine Translation (WMT) uses a mix of non-professional and professional translators to create the test sets for its annual 
shared translation tasks \cite{Callison-Burch2008a,WMT09-Findings}. 
For WMT09, the total cost of creating the test sets consisting of roughly 80,000 words across 3027 sentences in seven European languages was approximately \$39,800 USD, or 
slightly more than \$0.08 USD/word.  
For WMT08, creating test sets consisting of 2,051 sentences in six languages was approximately \$26,500 USD or slightly more than \$0.10 USD/word.

In this paper we examine the use of Amazon's Mechanical Turk (MTurk) to create translation test sets for statistical machine translation research.  
\newcite{Snow2008} showed that MTurk can be useful for creating data for a variety of NLP tasks, and that a combination of judgments from non-experts can attain expert-level quality in many cases. 
\newcite{callisonburch2009} showed that MTurk could be used for low-cost manual evaluation of  machine translation quality, and suggested that it might be possible to use MTurk to create MT 
test sets after an initial pilot study where turkers (the people who complete the work assignments posted on MTurk) produced translations of 50 sentences in five languages.   

This paper explores this in more detail by asking turkers to translate the Urdu sentences of the Urdu-English test set used in the 2009 NIST Machine Translation Evaluation Workshop. 
We evaluate multiple MT systems on both the professionally-produced NIST2009 test set and our MTurk-produced test set and find that the MTurk-produced test set yields essentially the same conclusions 
about system performance as the NIST2009 set yields. 

\section{Gathering the Translations via Mechanical Turk} \label{gathering}

The NIST2009 Urdu-English test set\footnote{http://www.itl.nist.gov/iad/894.01/tests/mt/2009/\\ResultsRelease/currentUrdu.html} 
is a professionally produced machine translation evaluation set, containing four human-produced reference translations for each of 
1792 Urdu sentences. 
We posted the 1792 Urdu sentences on MTurk and asked for translations into English.  
We charged \$0.10 USD per translation, giving us a total translation cost of \$179.20 USD.
A challenge we encountered during this data collection was that many turkers would cheat, giving us fake translations. 
We noticed that many turkers were pasting the Urdu into an online machine translation system and giving us the output as their
response even though our instructions said not to do this. 
We manually monitored for this and rejected these responses and blocked these workers from computing any of our future work assignments. 
In the future, we plan to combat this in a more principled manner by converting our Urdu sentences into an image and posting the
images. This way, the cheating turkers will not be able to cut and paste into a machine translation system. 

We also noticed that many of the translations had simple mistakes such as misspellings and typos. 
We wanted to investigate whether these would decrease the value of our test set so we did a second phase of data collection where we
posted the translations we gathered and asked turkers (likely to be completely different people than the ones who provided the 
initial translations) to correct simple grammar mistakes, misspellings, and typos. 
For this post-editing phase, we paid \$0.25 USD per ten sentences, giving a total post-editing cost of \$44.80 USD. 

In summary, we built two sets of reference translations, one with no editing, and one with post-editing. 
In the next section, we present the results of experiments that test how effective these test sets are for 
evaluating MT systems. 

\section{Experimental Results} \label{results}

A main purpose of an MT test set is to evaluate various MT systems' performances relative to each
other and assist in drawing conclusions about the relative quality of the translations produced by the
systems.\footnote{Another useful purpose would be to get some absolute sense of the quality of the translations but that
seems out of reach currently as the values of BLEU scores (the defacto standard evaluation metric) are difficult to map to
precise levels of translation quality.}
Therefore, if a given system, say System A, outperforms another given system, say System B, on a
high-quality professionally-produced test set, then we would want to see that System A also outperforms System B on our
MTurk-produced test set. It is also desirable that the magnitudes of the differences in 
performance between systems also be maintained. 

In order to measure the differences in performance, using the differences in the absolute magnitudes of the BLEU scores will not work
well because the magnitudes of the BLEU scores are affected by many factors of the test set being used, such as the number of
reference translations per foreign sentence. 
For determining performance differences between systems and especially for comparing them {\em across different test sets}, 
we use percentage of baseline performance. 
To compute percentage of baseline performance, we designate one system as the baseline system and use percentage of that baseline
system's performance. 
For example, Table~\ref{t:percentPerformance} shows both absolute BLEU scores and percentage performance for three MT systems when
tested on five different test sets. The first test set in the table is the NIST-2009 set with all four reference translations per
Urdu sentence. The next four test sets use only a single reference translation per Urdu sentence (ref 1 uses the first reference
translation only, ref 2 the second only, etc.). Note that the BLEU scores for the single-reference translation test sets are much 
lower than for the test set with all four reference translations and the difference in the absolute magnitudes of the BLEU scores
between the three different systems are different for the different test sets. However, the percentage performance of the MT systems
is maintained (both the ordering of the systems and the amount of the difference between them) across the different test sets. 

\begin{table}[t]
\begin{center}
\begin{tabular}{|l|c|c|c|} \hline
{\bf Eval} & {\bf ISI}      & {\bf JHU}      & {\bf Joshua}  \\
{\bf Set}  & {\bf (Syntax)} & {\bf (Syntax)} & {\bf (Hier.)} \\ \hline
NIST-2009  & 33.10          & 32.77          & 26.65               \\ \cline{2-4}
(4 refs)   & 100\%          & 99.00\%        & 80.51\%             \\ \hline
NIST-2009  & 17.22          & 16.98          & 14.25               \\ \cline{2-4}
(ref 1)    & 100\%          & 98.61\%        & 82.75\%             \\ \hline
NIST-2009  & 17.76          & 17.14          & 14.69               \\ \cline{2-4}
(ref 2)    & 100\%          & 96.51\%        & 82.71\%             \\ \hline
NIST-2009  & 16.94          & 16.54          & 13.80               \\ \cline{2-4}
(ref 3)    & 100\%          & 97.64\%        & 81.46\%             \\ \hline
NIST-2009  & 13.63          & 13.67          & 11.05               \\ \cline{2-4}
(ref 4)    & 100\%          & 100.29\%       & 81.07\%             \\ \hline
\end{tabular}
\end{center}
\caption{\label{t:percentPerformance} This table shows three MT systems evaluated on five different test sets. 
For each system-test set pair, two numbers are displayed. The
top number is the BLEU score for that system when using that test set. For example, ISI-Syntax tested on the NIST-2009
test set has a BLEU score of 33.10. The bottom number is the percentage of baseline system performance that is achieved. 
ISI-Syntax (the highest-performing system on NIST2009 to our knowledge) is used as the baseline. Thus, it will always have
100\% as the percentage performance for all of the test sets. 
To illustrate computing the percentage performance for the other systems,
consider for JHU-Syntax tested on NIST2009, that its BLEU score of 32.77 divided by the BLEU score of the baseline system is
$32.77 / 33.10 \approx 99.00\%$}
\end{table}

We evaluated three different MT systems on the NIST2009 test set and on our two MTurk-produced 
test sets
(MTurk-NoEditing and MTurk-Edited). Two of the MT systems (ISI Syntax \cite{Galley2004,Galley2006} and JHU Syntax \cite{Joshua-WMT} augmented with \cite{samt2006}) were 
chosen because they represent
state-of-the-art performance, having achieved the highest scores on NIST2009 to our knowledge. They also have very similar
performance on NIST2009 so we want to see if that similar performance is maintained as we evaluate on our MTurk-produced test
sets. The third MT system (Joshua-Hierarchical) \cite{Joshua-WMT}, an open source implementation of \cite{Chiang2007}, was chosen because though it is a competitive system, it had clear, markedly
lower performance on NIST2009 than the other two systems and we want to see if that difference in performance is also
maintained if we were to shift evaluation to our MTurk-produced test sets. 

Table~\ref{t:systemRelations} shows the results. There are a number of observations to make.
One is that the absolute magnitude of the BLEU scores is much lower for all systems on the MTurk-produced test sets than on the NIST2009 test
set. This is primarily because the NIST2009 set 
had four translations per foreign sentence whereas the MTurk-produced sets only have one translation per foreign sentence. 
Due to this different scale of BLEU scores, we compare performances using percentage of baseline performance. 
We use the ISI Syntax system as the baseline since it achieved the highest results on NIST2009.
The main observation of the results in Table~\ref{t:systemRelations} is that both the relative performance of the various MT systems
and the amount of the differences in performance (in terms of percentage performance of the baseline) are maintained when we use the
MTurk-produced test sets as when we use the NIST2009 test set. In particular, we can see that whether using the NIST2009 test set or
the MTurk-produced test sets, one would conclude that ISI Syntax and JHU Syntax perform about the same and Joshua-Hierarchical delivers
about 80\% of the performance of the two syntax systems. The post-edited test set did not yield different conclusions than the
non-edited test set yielded so the value of post-editing for test set creation remains an open question. 

\begin{table}
\begin{center}
\begin{tabular}{|l|c|c|c|} \hline
{\bf Eval} & {\bf ISI}      & {\bf JHU}      & {\bf Joshua}  \\
{\bf Set}  & {\bf (Syntax)} & {\bf (Syntax)} & {\bf (Hier.)} \\ \hline
NIST-      & 33.10          & 32.77          & 26.65               \\ \cline{2-4}
2009       & 100\%          & 99.00\%        & 80.51\%             \\ \hline
MTurk-       & 13.81          & 13.93          & 11.10               \\ \cline{2-4}
NoEditing  & 100\%          & 100.87\%       & 80.38\%             \\ \hline
MTurk-       & 14.16          & 14.23          & 11.68               \\ \cline{2-4}
Edited     & 100\%          & 100.49\%       & 82.49\%             \\ \hline
\end{tabular}
\end{center}
\caption{\label{t:systemRelations} This table shows three MT systems evaluated using the official NIST2009 test set and the
two test sets we constructed (MTurk-NoEditing and MTurk-Edited). For each system-test set pair, two numbers are displayed. The
top number is the BLEU score for that system when using that test set. For example, ISI-Syntax tested on the NIST-2009
test set has a BLEU score of 33.10. The bottom number is the percentage of baseline system performance that is achieved. 
ISI-Syntax (the highest-performing system on NIST2009 to our knowledge) is used as the baseline.} 
\end{table}

\section{Conclusions and Future Work} \label{conclusions}

In conclusion, we have shown that it is feasible to use MTurk to build MT evaluation sets at a significantly reduced cost. 
But the large cost savings does not hamper the utility of the test set for evaluating systems' translation quality. 
In experiments, MTurk-produced test sets lead to essentially the same conclusions about multiple MT systems' translation quality as much more expensive professionally-produced MT test sets.

It's important to be able to build MT test sets quickly and cheaply because we need new ones for new domains (as discussed in Section~\ref{introduction}). 
Now that we have shown the feasibility of using MTurk to build MT test sets, in the future we plan to build new MT test sets for specific domains (e.g., entertainment, science, etc.) and release them to
the community to spur work on domain-adaptation for MT.

We also envision using MTurk to collect additional training data to tune an MT system for a new domain. 
It's been shown that active learning can be used to reduce training data annotation burdens for a variety of NLP tasks (see, e.g., \cite{bloodgood2009a}). 
Therefore, in future work, we plan to use MTurk combined with an active learning approach to gather new data in the new domain to investigate improving MT performance for specialized domains.
But we'll need new test sets in the specialized domains to be able to evaluate the effectiveness of this line of research and therefore, we will need to be able to build new test sets. 
In light of the findings we presented in this paper, it seems we can build those test sets using MTurk for relatively low costs without sacrificing much in their utility for evaluating MT
systems. 

\section*{Acknowledgements}
This research was supported by the EuroMatrixPlus project funded by the European Commission, by the DARPA GALE program under Contract No. HR0011-06-2-0001, and the NSF under grant IIS-0713448.  Thanks to Amazon Mechanical Turk for providing a \$100 credit.

\bibliographystyle{naaclhlt2010}
\bibliography{paper}

\end{document}